\def\year{2021}\relax
\documentclass[letterpaper]{article} 
\usepackage{aaai21}  
\usepackage{times}  
\usepackage{helvet} 
\usepackage{courier}  
\usepackage[hyphens]{url}  
\usepackage{graphicx} 
\usepackage{graphicx}  
\usepackage{natbib}  
\usepackage{caption} 
\usepackage{amsfonts} 
\usepackage{amsmath}
\usepackage{url}
\usepackage[hidelinks]{hyperref}
\usepackage[utf8]{inputenc}

\usepackage{booktabs}
\usepackage{multirow}
\usepackage{subfig}
\usepackage{algorithm}
\usepackage{algorithmic}

\title{Generative Feature Replay with Orthogonal Weight Modification for Continual Learning}

\author{Gehui Shen,  Song Zhang, Xiang Chen, Zhi-Hong Deng\thanks{Corresponding author} \\}

\affiliations{
	    School of Electronics Engineering and Computer Science, Peking University\\
	    \{jueliangguke, songz, caspar, zhdeng\}@pku.edu.cn}

\newcommand{\tabincell}[2]{\begin{tabular}{@{}#1@{}}#2\end{tabular}} 

\DeclareMathOperator{\E}{\mathbb{E}}
\DeclareMathOperator{\I}{\mathbb{I}}
\begin{document}

\maketitle

\begin{abstract}
The ability of intelligent agents to learn and remember multiple tasks sequentially is crucial to achieving artificial general intelligence. Many continual learning (CL) methods have been proposed to overcome catastrophic forgetting which results from non i.i.d data in the sequential learning of neural networks. In this paper we focus on class incremental learning, a challenging CL scenario. For this scenario, generative replay is a promising strategy which generates and replays pseudo data for previous tasks to alleviate catastrophic forgetting. However, it is hard to train a generative model continually for relatively complex data. Based on recently proposed orthogonal weight modification (OWM) algorithm which can approximately keep previously learned feature invariant when learning new tasks, we propose to 1) replay penultimate layer feature with a generative model; 2) leverage a self-supervised auxiliary task to further enhance the stability of feature. Empirical results on several datasets show our method always achieves substantial improvement over powerful OWM while conventional generative replay always results in a negative effect. Meanwhile our method beats several strong baselines including one based on real data storage. In addition, we conduct experiments to study why our method is effective.
\end{abstract}

\section{Introduction}
Deep learning has achieved remarkable levels of performance for AI, exceeding the abilities of human experts on several particular tasks~\citep{prelu,alphago}. However, neural networks (NN) are prone to suffer from catastrophic forgetting~\citep{1989cataf,1999cataf} when learning multiple tasks in a sequential manner. This phenomenon results from the interference between the knowledge of previous tasks and current task because the data of previous tasks are unavailable and leads to significant degradation of previous tasks' performance. In contrast, humans excel at learning new skills and accumulating knowledge continually throughout their lifespan. Continual learning (CL)~\citep{review} aims to bridge this gap and has became an important challenge of AI research. It allows intelligent agents to reuse and transfer old knowledge meanwhile meets the real-world situation where training data are hardly available simultaneously.

According to whether task identity is provided and whether it must be inferred during test, there are mainly three CL scenarios~\citep{3sc,3sc2} i.e. task incremental learning, domain incremental learning and class incremental learning (CIL). CIL is the most challenging scenario in which the classes of each task are disjoint and the model is trained to distinguish classes of all tasks with a shared output layer a.k.a. ``single-head''. In this paper we focus on CIL. CIL corresponds to the problem of learning new classes of objects incrementally which is common in real-world applications. As the data of old classes are unavailable, the shared output layer exacerbates the forgetting of previous tasks. Some CL methods~\citep{ewc,lwf} performing well in the other two scenarios almost totally get failed in CIL~\citep{3sc,3sc2}. A naive approach to alleviate this problem is to store a subset of data of previous tasks and replay them to classifier when learning new tasks~\citep{icarl,gem}. However it violates the main protocol in CL that only data of current task are available. Data privacy and security concerns also question the practical value of real data replay methods. In this paper, we aim to design CIL method without any data storage.

Inspired by complementary learning systems (CLS) theory~\citep{cls} about biological mechanisms, generative replay (GR)~\citep{dgr} approach has made some progress in CIL. Instead of storing the real data, a generative model, such as generative adversarial network (GAN)~\citep{gan} or VAE~\citep{vae}, is trained to learn the data distribution of previous tasks. During training, both the real data of the current task and the synthesized data sampled from the generator are fed into classifier. As the synthesized data represent the distribution of old classes approximately, classifier can retain the knowledge of previous tasks. While learning the new knowledge, the generator are trained in the same manner to prevent catastrophic forgetting. Despite GR works well on simple datasets such as MNIST~\citep{mnist}, it is far from solving CIL completely. As pointed out in~\citet{19ijcnn}, training a GAN in the CL scenario is difficult because of catastrophic forgetting. For example, generative replay performs poorly when being applied on CIFAR10~\citep{cifar}, a real-world image dataset~\citep{19ijcnn}. \citet{pgma} also find it is impractical to replay text data with a generative model.

Recently, orthogonal weight modification (OWM)~\citep{owm} algorithm has been proposed which is applicable in the above three CL scenarios and can be considered as a state-of-the-art method for CIL. The main idea behind OWM algorithm is that to protect previously learned knowledge, we update the parameters only in the direction orthogonal to the subspace spanned by all previous inputs fed into the network. In this way, during training new inputs, the network can keep the learned input-output mappings invariable. 

In this paper, on the basis of OWM algorithm, we propose to generate and replay feature instead of raw data to improve the performance of CIL. Specifically, we utilize OWM algorithm on classifier meanwhile we train a generator to learn the distribution of the penultimate layer feature. When training the subsequent tasks, generated pseudo features together with the penultimate layer features of current data are fed into the last fully connected (FC) layer. Thanks to the effect of OWM, the features are stable which makes the feature replay feasible potentially. 

The motivation is three-fold: 1) Although OWM can keep the ability of distinguishing the classes within one task, data from classes that belong to different tasks are never fed into classifier simultaneously. Therefore classifier is prone to confuse about task identity when classifying over all classes. Replaying features of previous tasks can alleviate this problem. 2) As raw data contain many details not related to class information, learning to continually generate real-world data is hard. However, the distribution of high-level features are relatively simple, which lessens the difficulty of training the generator. 3) Almost all NN-based classifiers' output layers are fully connected so that generating the penultimate layer feature is a universal strategy.

To further enhance the stability of features, we also leverage a self-supervised learning~(SSL) objective to introduce an auxiliary task. SSL recently has shown powerful ability in unsupervised visual representation learning~\citep{infonce,moco,clr}, which can be used to extract general features for downstream supervised tasks. More general features can implicitly reduce the change of classifier when adapting features to new tasks. In this way, we can expect the penultimate layer feature to be more stable thus feature replay is more reliable.

In summary, our contributions are as follows: 
\begin{enumerate}
\item Based on the hypothesis that the function of OWM facilitates the success of generative feature replay (GFR), we propose a hybrid CIL approach which aims to improve OWM method with GFR. 
\item To enhance the stability of the feature further, we propose to combine our method with an SSL auxiliary task to strengthen generative feature replay.
\item Experimental results on several real-world image and text datasets show the superiority of proposed method to a series of strong CIL methods, including OWM baseline. We also conduct exploration experiments to demonstrate OWM is critical to GFR and why proposed method can improve OWM.
\end{enumerate}

\section{Related Work}\label{sec2}
The study of catastrophic forgetting in neural networks originated in the 1980s~\citep{1989cataf,reh93}. Under the revival of neural networks, overcoming catastrophic forgetting in continual learning setting has drawn much attention again~\citep{14r,progress,ewc}. In this section, we mainly review recent continual learning literature which is closely related to our work. 

Contemporary continual learning strategies can be divided roughly into four categories which are \textit{Regularization}, \textit{Task-specific}, \textit{Replay} and \textit{Gradient Projection} respectively. It should be noted that some CL works may incorporate more than one strategy. 

The most famous \textit{Regularization} method is EWC~\citep{ewc} which protects parameters by adding L2 penalty term based on the importance of each parameter to previous tasks. Some works focus on improving the estimate of importance~\citep{si,rwalk,mas}. \citet{lwf} propose another type of method which encourages the predicted probabilities on old classes of current classifier to approximate that of the old classifier. Such regularization methods are effective on task/domain incremental learning scenarios, however when applied to CIL scenario, they almost totally fail~\citep{3sc,3sc2}.
 
\textit{Task-specific} methods prevent knowledge interference by establishing task-specific modules for different tasks. The task-specific modules can be hidden units~\citep{xdg}, network weights~\citep{packnet} and dynamically growing sub-networks~\citep{progress}. This type of strategy is designed for task incremental learning. During test task identity is necessary therefore they are not applicable for CIL.

\textit{Replay} strategy is initially proposed to relearn a subset of previously learned data when learning the current task~\citep{reh93}. Some recent works~\citep{icarl,gem,eeil,bic} storing a subset of old data fall into this category which we call \textit{real data replay}. Real data replay not only violates the CL requirement that old data are unavailable, but also is against the notion of bio-inspired design. According to CLS theory~\citep{cls} the hippocampus encodes and replays recent experiences to help the memory in the neocortex consolidate. Some evidence illustrates hippocampus works like a generative model than a replay buffer, which has inspired the proposal of generative replay (GR)~\citep{dgr,mergan,dgm}. GR utilizes GAN framework to train a generator to learn the distribution of old data. When learning new tasks the pseudo data generated by the generator are replayed to classifier. Due to the power of approximating distribution of GAN, replayed data reduce the shift of data distribution, especially in CIL scenario, thus can alleviate catastrophic forgetting. As the generator is also trained continually with replayed data, GR may break down when it encounters complex data~\citep{19ijcnn}. A remedy is to encode the raw data into features with a feature extractor pre-trained on large-scale datasets and take features as inputs for training and replay~\citep{pgma,ptfr}. However, such a pre-trained model is not often easily obtained. In contrast, we firstly propose a successful method to replay features without pre-training. 

When we were working at this research, we found a concurrent work~\citep{fdgfr} which also proposes GFR and highly coincides with our idea. However, we need to emphasize that our GFR is based on OWM method and is the first work to improve the state-of-the-art OWM for CIL. \citet{fdgfr} exploit a feature distillation loss to keep the stability of feature so we call it FD+GFR while our method is named OWM+GFR. Experimental results will show OWM+GFR performs much better than FD+GFR and OWM is critical to the success of GFR. Our contributions also reflect in introducing SSL auxiliary task and evaluation on text datasets.

The last \textit{Gradient Projection} strategy retains previously learned knowledge by keeping the old input-output mappings NNs induce fixed. To meet this goal, gradients are projected to the subspace orthogonal to the subspace spanned by inputs of previous tasks. \citet{cab} and \citet{owm} respectively propose CAB and OWM algorithm which resort to different mathematical theories to compute the orthogonal subspace approximately. This strategy makes the features of previous tasks stable when learning new tasks, which allows us to conduct generative feature replay.

\begin{figure}[tb]
\centering
\subfloat[Training Generator $(G_n,D_n)$ on the $n$-th task.]{\label{fig1:a}\includegraphics[height=0.13\textheight]{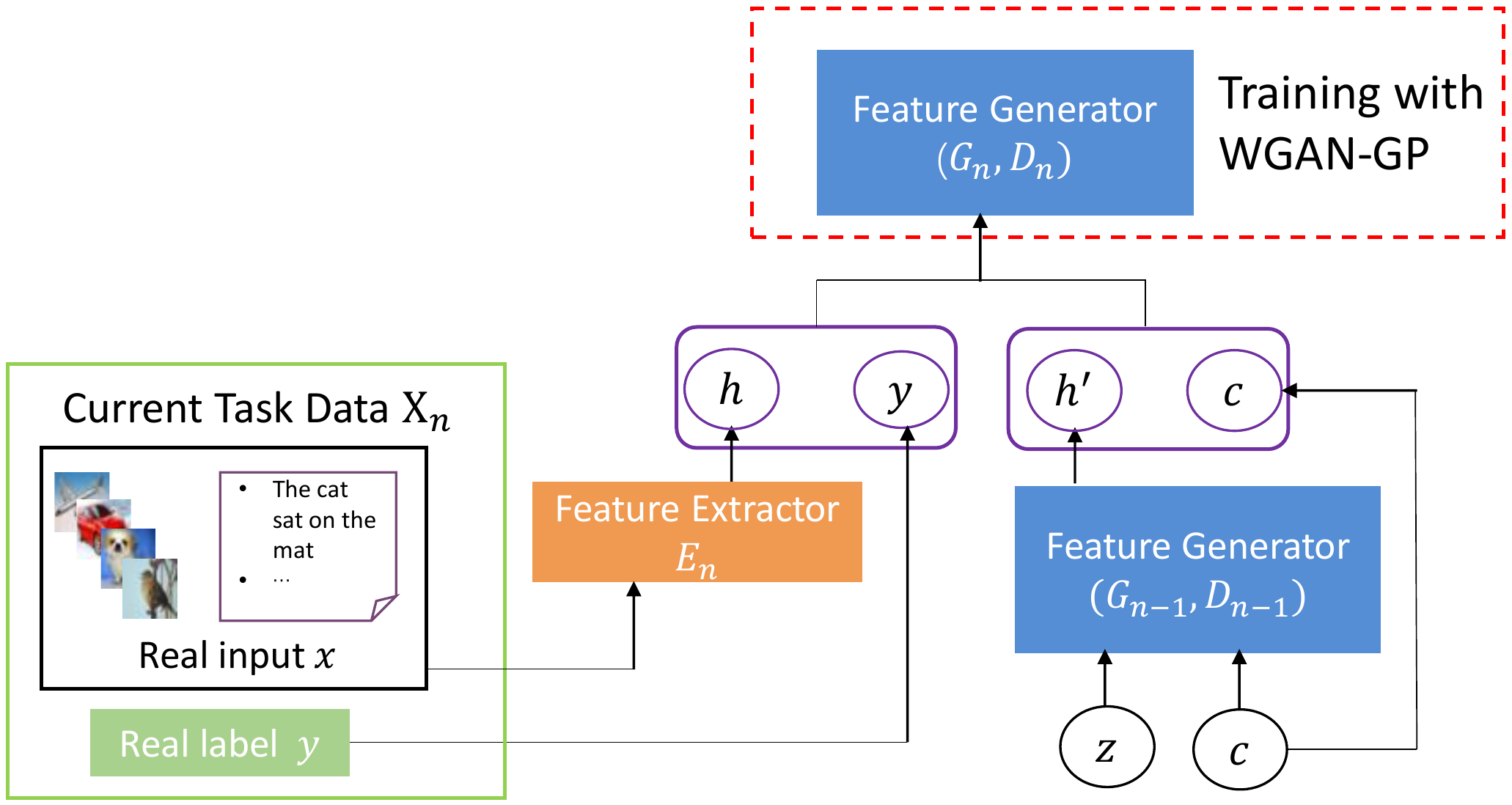}}
\\
\vspace{-1mm}
\subfloat[Training Classifier $C_{n+1}=(E_{n+1},F_{n+1})$ on the ($n$+1)-th task.]{\label{fig1:b}\includegraphics[height=0.13\textheight]{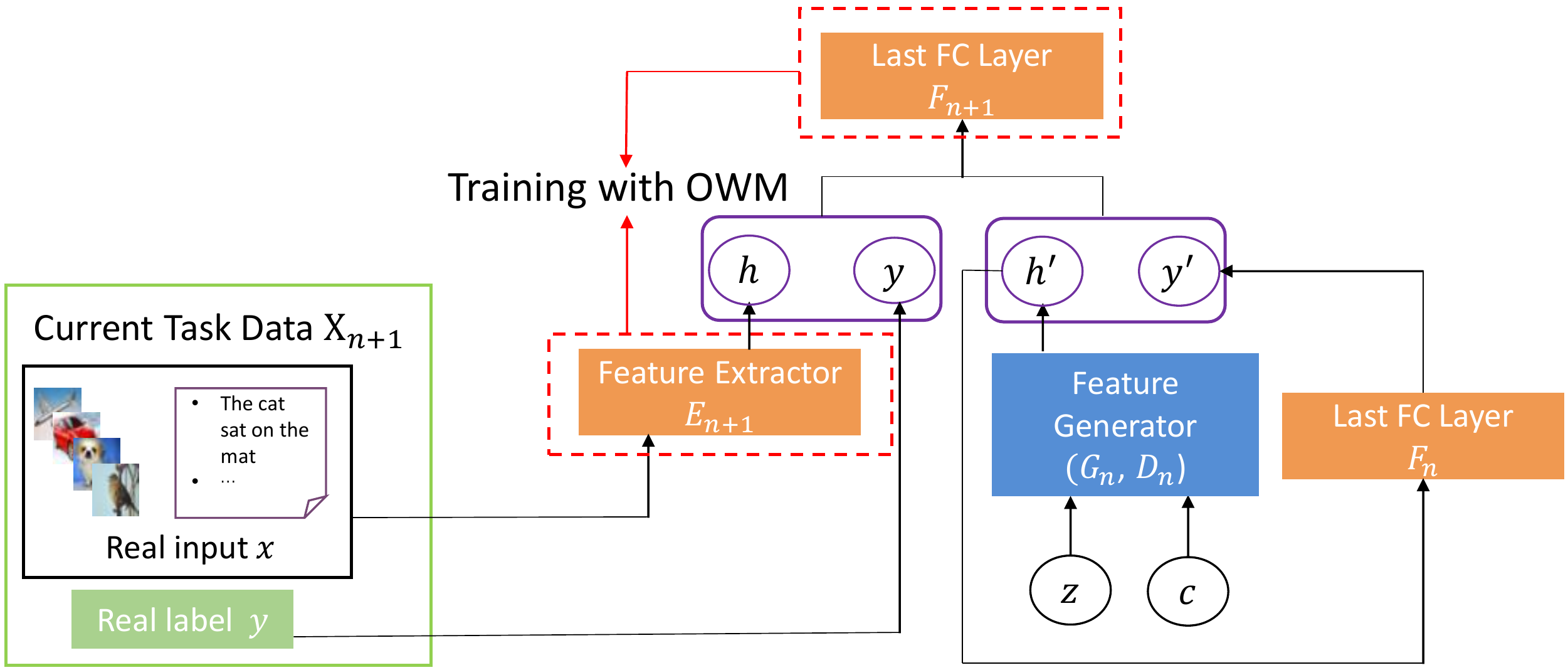}}
\caption{Model overview of proposed method. In each subfigure, the modules in red dashed box are trained and the others are fixed.}
\label{figure1}
\end{figure}

\section{Methodology}
In this section, we describe our continual learning framework which is a hybrid approach incorporating subspace method and generative replay strategy. As illustrated in Figure~\ref{figure1}, our framework is composed of a classifier $C$ and a GAN-based generative model $(G,D)$ where $G$ and $D$ are generator and discriminator respectively~\citep{gan}. We divide $C$ into two parts: the last FC layer $F$ and all previous layers $E$ which is treated as feature extractor. The way we train the generative model is similar with prior generative replay methods including DGR~\citep{dgr} and MeRGAN-JTR~\citep{mergan}. However, our framework is different from them in the following: 1) We train a separated classifier $C=(E,F)$ with OWM algorithm~\citep{owm} for classification. 2) In this way, we propose to train a GAN to generate the penultimate layer feature $\mathbf{h}$, the output of $E$, for replay to alleviate the catastrophic forgetting in the last FC layer $F$. 3) For image data which are widely used in CL field, we use a self-supervised auxiliary task to enhance the stability of $\mathbf{h}$ further.

\subsection{Training Classifier with OWM algorithm}

Conventionally, for an FC layer the weight matrix $W$ is updated by gradient descent algorithm and learning the $n$-th task $T_n$ leads to the change  $W^n=W^{n-1}-\lambda\Delta W^{n}_{\rm BP}$ where $\lambda$ is learning rate and $\Delta W^{n}_{\rm BP}$ represents the gradient computed by back propagation (BP) during training $T_n$. When testing on the previous tasks $T_{<n}$, the FC layer's output $\mathbf{x}^{<n,n}_{out}$ is deviated from the optimum value $\mathbf{x}_{out}^{<n,n-1}$ after learning $T_{n-1}$, i.e. $\mathbf{x}_{out}^{<n,n}=W^n \mathbf{x}_{in}^{<n}=W^{n-1}\mathbf{x}_{in}^{<n}-\lambda \Delta W^{n}_{\rm BP}\mathbf{x}_{in}^{<n}\not=W^{n-1} \mathbf{x}_{in}^{<n}=\mathbf{x}_{out}^{<n,n-1}$. The deviation accumulates across layers and causes catastrophic forgetting on previous tasks. To overcome this, \citet{owm} have developed OWM algorithm to project $\Delta W^{n}_{\rm BP}$ to the subspace orthogonal to the input of all previous tasks: $P_n=I-A_n(A_n^TA_n + \alpha I)^{-1}A_n^T$, where matrix $A_n = [\mathbf{x}_1,\cdots,\mathbf{x}_{n-1}]$ consists of input vectors which has been trained before as columns\footnote{Without loss of generality, here we treat $\mathbf{x}_i$ as a vector for brevity,  which means each task has only one input data.} and $\alpha$ is a small constant to resist in noise. The gradient is modified with the projector $P_n$: $\Delta W^{n}_{\rm OWM}=\Delta W^{n}_{\rm BP}P_n$. Because for any input $\mathbf{x}_{in}^{<n}$ in the input space $A_n$ we have $A_n(A_n^TA_n + \alpha I)^{-1}A_n^T\mathbf{x}_{in}^{<n}\approx\mathbf{x}_{in}^{<n}$ thus $P_n\mathbf{x}_{in}^{<n}\approx \mathbf{0}$, OWM can keep the output invariable approximately after gradient descent update:  
\begin{equation*}
\begin{split}
\mathbf{x}_{out}^{<n,n}&=W^n \mathbf{x}_{in}^{<n}\\
&=W^{n-1}\mathbf{x}_{in}-\lambda \Delta W^{n}_{\rm OWM}\mathbf{x}_{in}^{<n}\\
&=\mathbf{x}_{out}^{<n,n-1}-\lambda\Delta W^{n}_{\rm BP}P_n\mathbf{x}_{in}^{<n}\\
&\approx\mathbf{x}_{out}^{<n,n-1}
\end{split}
\end{equation*}

\noindent Such a property makes the network be capable of maintaining previously learned input-output mappings meanwhile the new tasks can be learned. The extra operation is computing projector $P$ and projected gradient $\Delta W_{\rm OWM}$ before running gradient descent. We can calculate $P$ in an efficient online manner as described in~\citep{owm}:
\begin{equation*}
P^k=P^{k-1}-(\alpha+{\bar{\mathbf{x}}_{k}^{T}}P^{k-1}\bar{\mathbf{x}}_{k}^{})^{-1}P^{k-1}{\bar{\mathbf{x}}_{k}^{}}\bar{\mathbf{x}}_{k}^{T}P^{k-1}
\end{equation*}
where $k$ indexes the mini-batch and ${\bar{\mathbf{x}}_{k}^{}}$ is the mean of the $k$-th mini-batch's inputs. Although we use FC layer to explain here, OWM algorithm can be applied to other NNs such as convolution neural networks (CNN). See the original paper~\citep{owm} for more details about OWM.  

\subsection{Generative Feature Replay}
Although OWM is a state-of-the-art method in CIL scenarios, projected gradients can mainly protect the knowledge of distinguishing the classes in the same task. The classifier potentially has confusion about the task identity during test as the data from different tasks are never trained together, which is the major difficulty in CIL. 

To alleviate this problem, we propose Generative Feature Replay (GFR), which replays the penultimate layer feature $\mathbf{h}$ instead of the raw data to improve OWM. We use an Auxiliary Conditional GAN (AC-GAN)~\citep{acgan} as generator, which allows conditional generation with labels and has been proven to work better than unconditional GAN for continual learning~\citep{mergan}. To avoid confusion, we call generative replay like DGR~\citep{dgr,mergan} for the original input as Generative Input Replay (GIR).

In our framework, the whole model is comprised of a classifier $C({\theta^C})$, a generator $G({\theta^G})$ and a discriminator $D({\theta^D})$. The discriminator gives two outputs: $D^{dis}$ means the probability that the input is real, like vanilla GANs and $D^{cls}$ predicts the input's class label. The generator takes a noise $z$ and a class label $c$ as inputs. We further divide $C({\theta^C})$ into two parts: the last FC layer $F({\theta^F})$ and feature extractor $E({\theta^E})$. As $C$ is trained with OWM, given the inputs of previous tasks, each layer's outputs, including $\mathbf{h}$ remain stable when learning new tasks continually. Thus we can employ the AC-GAN to generate $\mathbf{h}$ for replay, which can reduce the difficulty of training the generator compare with GIR. 

We implement AC-GAN  with WGAN-GP technique~\citep{wgangp} for more stable training. The loss functions which $G({\theta^G_n})$ and $D({\theta^D_n})$ are trained to minimize when training the $n$-th task are as follows:
\begin{equation*}
\mathcal{L}(\theta^G_n) = -\mathcal{L}_{GAN}(\theta^G_n) + \mathcal{L}_{AC}(\theta^G_n)
\end{equation*}
\begin{equation*}
\mathcal{L}(\theta^D_n) = \mathcal{L}_{GAN}(\theta^D_n) + \mathcal{L}_{AC}(\theta^D_n) + \lambda_{GP}\mathcal{L}_{GP}
\end{equation*}
$\mathcal{L}_{GAN}$ is from $D^{dis}$ to discriminate real and fake features:
\begin{equation*}
\begin{split}
\mathcal{L}_{GAN} = &-\E_{(x,y)\sim X_n}[D^{dis}(E(x;\theta^E_n);\theta^D_n)] \\ 
&- \E_{z\sim p_z,c\sim p_{c_{1:n-1}}}[D^{dis}(G(z,c;\theta^G_{n-1});\theta^D_n)] \\
&+ \E_{z\sim p_z,c\sim p_{c_{n}}}[D^{dis}(G(z,c;\theta^G_{n});\theta^D_n)]
\end{split}
\end{equation*}
where $X_n$ is the real data distribution of $n$-th task. $p_z$ is the noise distribution. $p_{c_{1:n-1}}$ and $p_{c_{n}}$ represent label distribution for the first $n-1$ tasks and the $n$-th task respectively. The first term refers to real features of data in $T_n$ while the third term corresponds to fake features generated by the current generator $G(\theta^G_n)$. To allow $G(\theta^G_n)$ to be able to generate features of previous tasks, the fake features generated by the old generator $G(\theta^G_{n-1})$ are also considered as real features which corresponds to the second term. Similarly, the $\mathcal{L}_{AC}$ also has three parts:
\begin{equation*}
\begin{split}
\mathcal{L}_{AC} &= \E_{(x,y)\sim X_n}[\mathcal{L}_{CE}(D^{cls}(E(x;\theta^E_n);\theta^D_n),y)] \\
&+ \E_{z\sim p_z,c\sim p_{c_{1:n-1}}}[\mathcal{L}_{CE}(D^{cls}(G(z,c;\theta^G_{n-1});\theta^D_n),c)] \\
&+ \E_{z\sim p_z,c\sim p_{c_{n}}}[\mathcal{L}_{CE}(D^{cls}(G(z,c;\theta^G_{n});\theta^D_n),c)]
\end{split}
\end{equation*}
where $\mathcal{L}_{CE}$ means the cross-entropy loss. Here discriminator and generator are both trained to minimize the classification loss for real and fake features. $\mathcal{L}_{GP}$ is gradient penalty term ~\citep{wgangp}. It should be noted that when training $(\theta^G_n,\theta^D_n)$, $\theta^C_n=(\theta^E_n,\theta^F_n)$ and $\theta^G_{n-1}$ are fixed. 

Generator $G(\theta^G_n)$ is used to generate replayed features of the first $n$ tasks $\widetilde{\mathbf{h}}_n=G(z,c;\theta^G_{n})$ when training classifier in $T_{n+1}$. The loss function of classifier is as follows:
\begin{equation*}\label{eq1}
\begin{split}
&\mathcal{L}(\theta_{n+1}^E,\theta_{n+1}^F)=\mathcal{L}_{CLS}(\theta_{n+1}^E,\theta_{n+1}^F) + \mathcal{L}_{GFR}(\theta_{n+1}^F)\\
&\mathcal{L}_{CLS}=\E_{(x,y)\sim X_{n+1}}[\mathcal{L}_{CE}(C(x;\theta_{n+1}^E,\theta_{n+1}^F),y)]\\
&\mathcal{L}_{GFR}=\E_{z\sim p_z,c\sim p_{c_{1:n}}}[\mathcal{L}_{DT}(F(\widetilde{\mathbf{h}}_n;\theta_{n+1}^F),F(\widetilde{\mathbf{h}}_n;\theta_{n}^F))]
\end{split}
\end{equation*}
where the two terms correspond to the loss of true data and replayed features $\widetilde{\mathbf{h}}_n$ respectively. In the second term, $\mathcal{L}_{DT}$ represents distillation loss~\citep{dk} which is widely-used in CL literature~\citep{lwf,eeil} for transferring learned knowledge better. The probabilities predicted by the old classifier $F({\theta^F_{n}})$ are regraded as soft labels of replayed features. We set the temperature ${\rm T}$ to 2 for all experiments.

\subsection{Self-Supervised Learning Auxiliary Task}
Though OWM can maintain previously learned feature ideally, the mathematical approximations when computing orthogonal subspaces for practical consideration limit the actual effect of OWM (otherwise it can solve CL perfectly). To further improve GFR with OWM, we attempt to reduce the change of feature $\mathbf{h}$ from another perspective. Our motivation is \textit{if the model can extract more general features during training instead of merely focusing on the features for solving the current task, the features will need less change when learning subsequent tasks}.  

Recently, self-supervised learning has shown powerful ability for unsupervised representation learning, e.g. BERT~\citep{bert} in NLP, MoCo~\citep{moco} in CV. A series of self-supervised proxy objectives are designed~\citep{jigsaw,rgb,rotate,infonce}. In this way, informative features can be learned from large scale dataset without supervised signal and effectively benefit to downstream supervised tasks. Inspired by the progress in this field, we aim to use a self-supervised proxy objective to induce the encoder $E$ to learn more general features which are not only beneficial for current task. We expect to this type of inductive bias can meet the above motivation.

As the first step of exploration, we choose predicting rotation~\citep{rotate} SSL objective for image data. We firstly define a set of $K$ discrete rotation transformations $R=\{g_k(\cdot)\}_{k=1}^K$ where $g_1(\cdot)$ is the identity transformation. We add a new FC layer $F_r$ on the top of $E$ to predict which type of transformation is conducted given an input data $x^k=g_k(x)$ where $x$ is the original image. Formally, for task $T_n$, the SSL objective is as follows:
\begin{equation*}
\mathcal{L}_{SSL}=\E_{(x,y)\sim X_n,g_k(\cdot)\sim R}[\mathcal{L}_{CE}
(F_r(E(g_k(x))),k)]
\end{equation*}

For each mini-batch data, we first choose a random rotation transformation $g_k(\cdot)$ and then training the augmented classifier $C'=(E,F,F^r)$ with the following loss:
\begin{equation*}
\mathcal{L}(\theta_{n}^E,\theta_{n}^F,\theta_{n}^{F^r})=\mathcal{L}_{CLS}+\mathcal{L}_{GFR} + \alpha \mathcal{L}_{SSL}
\end{equation*}

\noindent where the $\mathcal{L}_{CLS}$ term should be revised correspondingly:
\begin{equation*}
\mathcal{L}_{CLS}=\E_{(x,y)\sim X_{n},g_k(\cdot)\sim R}[\mathcal{L}_{CE}(C(g_k(x);\theta_{n}^E,\theta_{n}^F),y)]
\end{equation*}

In this way, training $C$ is in a multi-task learning manner. The SSL objective is an auxiliary task to facilitate the feature $\mathbf{h}$ to extract global information. Thus $\mathbf{h}$ tends to be more stable to benefit to GFR thanks to the ability of OWM. However, for text data, it is hard to find a similar SSL objective, which we leave for the future work.

According to preliminary experiments, we found that introducing the transformed data into $\mathcal{L}_{CLS}$ will increase the difficulty of training classifier. To prevent the degradation of performance, during test we feed $K$ transformed images into $E$ and use the average of $K$ features $\mathbf{h}$ for final classification. When training generator, we only use the original images because using transformed images has no effect on the result, but increases the training time.

\subsection{Discussions}
In fact, our feature replay strategy only has effects on the last FC layer which is ubiquitous in NN-based classifiers. Thus for different types of input data and NN models, GFR can be applied universally without any special modifications. However, GIR needs to design different types of generator for different data. For example, a CNN-based GAN cannot be used to generate text data. Some recent works~\citep{bic,reb,il2m,wa} also improve CIL performance by eliminating the bias in the last FC layer. However they all depend on real data replay and are perpendicular to our work.

\begin{table}[t]
\footnotesize
\begin{center}
\begin{tabular}{ccccc}

\toprule[1pt]
\bf Dataset& \textbf{Type} & \#\textbf{Train}/\#\textbf{Test} & \#\textbf{Class} & \#\textbf{Task}\\ \hline \specialrule{0em}{1.3pt}{1.3pt}
SVHN& Image&73257/26032& 10& 5 \\
CIFAR10& Image&50000/10000& 10& 5\\
CIFAR100& Image& 50000/10000& 100&2/5/10/20\\
THUCNews& Text&50000/15000& 10& 5\\
DBPedia& Text& 560000/70000&14&5 \\
\toprule[1pt]
\end{tabular}
\end{center}
\caption{Details about five datasets.}
\label{table1}
\end{table}

\begin{table*}[tb]
\scriptsize
\begin{center}
\begin{tabular}{cc|c|cccc|c|c} 
 \hline 

Methods & \tabincell{c}{SVHN \\ (5 tasks)} &\tabincell{c}{CIFAR10 \\ (5 tasks)} &\tabincell{c}{CIFAR100 \\ (2 tasks)} &\tabincell{c}{CIFAR100 \\ (5 tasks)}&\tabincell{c}{CIFAR100 \\ (10 tasks)}&\tabincell{c}{CIFAR100 \\ (20 tasks)}& \tabincell{c}{THUCNews \\ (5 tasks)} & \tabincell{c}{DBPedia\\(5 tasks)}\\
\hline 

EWC~\citep{ewc} &12.25$\pm$0.13&18.53$\pm$0.11&24.31$\pm$0.63&12.53$\pm$0.69&7.56$\pm$0.25&4.15$\pm$0.11&19.88$\pm$0.03&15.49$\pm$1.42\\
MAS~\citep{mas} &18.11$\pm$1.80&20.24$\pm$1.52&29.04$\pm$0.92&14.35$\pm$0.22&8.45$\pm$0.27&5.27$\pm$0.64&34.80$\pm$6.04&29.80$\pm$5.78\\
\hline

GIR &67.50$\pm$0.83&22.39$\pm$0.83&36.48$\pm$0.61&25.52$\pm$0.46&15.23$\pm$0.62&9.19$\pm$0.43&38.67$\pm$5.57&N/A\\
PGMA~\citep{pgma} &--&40.47&--&--&--&--&52.93&69.68 \\
DGM~\citep{dgm} &73.01$\pm$0.77&50.53$\pm$0.46&28.23$\pm$0.75&25.43$\pm$0.14&24.09$\pm$0.19&17.42$\pm$1.25&--&-- \\

\hline
OWM~\citep{owm} & 73.92$\pm$0.75 & 54.52$\pm$0.45& 42.28$\pm$0.35&34.16$\pm$0.43& 30.54$\pm$0.91&27.64$\pm$0.42&79.65$\pm$0.64&91.45$\pm$0.62\\
OWM+GIR & 73.65$\pm$0.94&52.58$\pm$1.06&\textbf{44.13$\pm$0.37}& 32.70$\pm$0.50&28.67$\pm$0.42&26.30$\pm$0.60&72.15$\pm$3.19&N/A\\

\hline
naive GFR & 12.43$\pm$0.03 & 18.95$\pm$0.03 &32.28$\pm$0.61&19.10$\pm$1.08&10.64$\pm$0.94&5.28$\pm$0.76&19.86$\pm$0.02&14.79$\pm$0.78\\
FD+GFR \cite{fdgfr} & 59.00$\pm$0.81 & 42.59$\pm$0.94 &34.33$\pm$0.70&22.01$\pm$0.44&13.81$\pm$0.92& 6.55$\pm$1.26 &35.92$\pm$1.20&66.34$\pm$0.97\\
OWM+GFR (Proposed) & 75.82$\pm$0.44$^\dagger$ & 56.07$\pm$0.67$^\dagger$ & 42.58$\pm$0.50  & 35.35$\pm$0.15$^\dagger$& 32.30$\pm$0.61$^\dagger$& 27.84$\pm$0.63 &\textbf{81.22$\pm$0.25}$^\dagger$&\textbf{92.57$\pm$0.28}$^\dagger$\\
OWM+SDA+GFR (Proposed) & \textbf{78.36$\pm$0.56}$^\dagger$ &  \textbf{57.81$\pm$0.74}$^\dagger$ & 43.51$\pm$0.30$^\dagger$ & \textbf{36.32$\pm$0.14}$^\dagger$ & \textbf{33.22$\pm$0.52}$^\dagger$ &\textbf{28.30$\pm$0.46}$^\star$ &--&--\\
\hline 
iCaRL (B=200)~\citep{icarl} &45.96$\pm$1.72&46.46$\pm$1.46&21.30$\pm$0.83&16.62$\pm$0.70&11.54$\pm$0.51&6.57$\pm$0.79&80.74$\pm$0.59&91.87$\pm$1.17\\ 
iCaRL (B=2000)~\citep{icarl} &67.91$\pm$0.84&57.66$\pm$0.86&36.09$\pm$1.34&30.07$\pm$0.98&19.53$\pm$1.02&10.37$\pm$0.65&87.26$\pm$1.12&96.03$\pm$0.38\\ 
\hline 
Joint Training (Upper Bound) &92.34&76.87&\multicolumn{4}{c|}{53.02}&96.54&98.77 \\
 \hline 
\end{tabular}	
\end{center}
\caption{Test accuracy after all tasks are learned. We report the mean and standard error over 5 runs with different seeds. For GIR, we use MeRGAN~\citep{mergan} method on image data and DGR~\citep{dgr} method on text data. We conduct $t$-test between proposed methods and OWM: $^\star$ indicates $p\text{-value}<0.05$ and $^{\dagger}$ indicates $p\text{-value}<0.01$.}
\label{table2}
\end{table*}

\begin{figure*}[ht]
\centering
\includegraphics[height=0.25\textwidth]{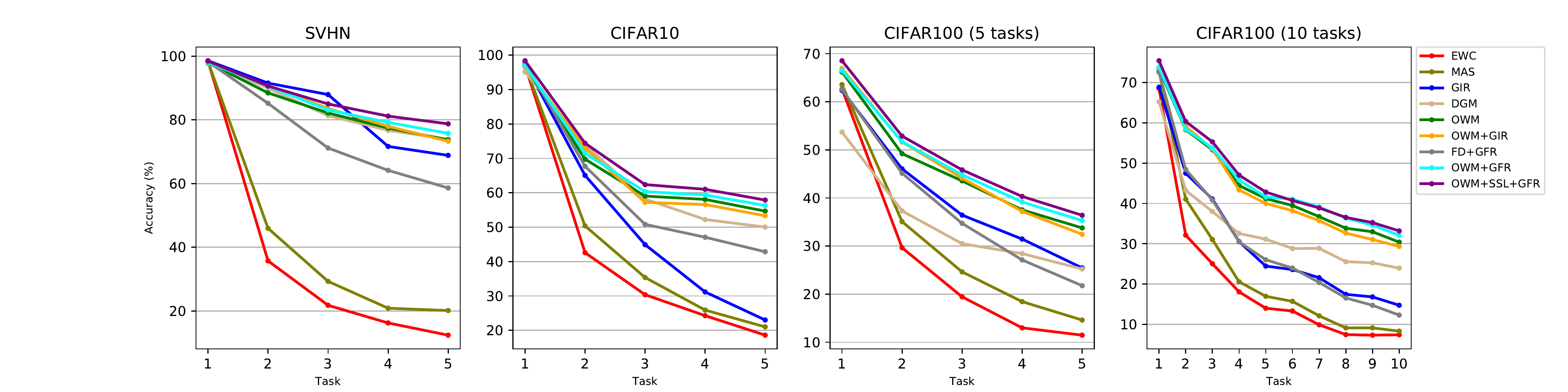}
\caption{Test accuracies on all classes of already learned tasks after each task is learned in 4 settings.}
\label{figure2}
\end{figure*}

\section{Experiments}
\subsection{Datasets, Baselines and Model Settings}
\subsubsection{Datasets.} 
To evaluate our method, we conduct experiment in CIL settings, where each task corresponds to a disjoint subset of classes of the whole dataset and the classifier only has one shared output layer. We use image datasets and two text datasets which are in detail in Table~\ref{table1}. The image datasets we use are collected from real world and challenging for training generative model in CL settings. We also use text datasets to reflect the versatility of proposed method. To make a fair comparison, we randomly select a subset of test data for validation and the left data are considered as test data following~\citet{pgma}. For CIL scenario, we split all datasets into 5 tasks and the number of class of each task is equal except DBPedia, where the 5 tasks have 3, 3, 3, 3, and 2 classes respectively. We also establish the settings of 2/10/20 tasks on CIFAR100 to further evaluate the performance under different numbers of tasks.

\subsubsection{Baselines.} 
For image data, we choose the following baselines for comparison: 1) two  regularization method: \textbf{EWC}~\citep{ewc} and \textbf{MAS}~\citep{mas}; 2) 3 recent GR methods: \textbf{MeRGAN}~\citep{mergan}, \textbf{DGM}~\citep{dgm} and \textbf{PGMA}~\citep{pgma}; 3) \textbf{OWM}~\citep{owm}, the state-of-the-art subspace method which is the basis of our method. We also include \textbf{OWM+GIR} which incorporates OWM method with GIR using MeRGAN. 4) For GFR, we include \textbf{naive GFR} which directly uses GFR without any other CL method and \textbf{FD+GFR}~\citep{fdgfr}. 5) We also compare with \textbf{iCaRL}~\citep{icarl}, a strong baseline with real data storage. We evaluate iCaRL with $B=200$ and $B=2000$ where $B$ is the size of storage data. Our method are \textbf{OWM+GFR} and \textbf{OWM+SSL+GFR} where the latter employs SSL auxiliary task. For text data, we only compare GFR methods with EWC, MAS, GIR, OWM, OWM+GIR and PGMA which first evaluated on those text datasets. The generator of GIR is SeqGAN~\citep{seqgan} and is trained in the DGR~\citep{dgr} framework.

\subsubsection{Model Settings.} 
We reimplement all baselines except PGMA and DGM\footnote{PGMA's results are taken from the original paper and DGM's results are obtained with source code~(https://github.com/SAP-samples/machine-learning-dgm).}. To make a fair comparison as much as possible, for images dataset we use classifier with the same architecture as in~\citep{owm} which has 3-layer CNN with 64, 128, 256 2$\times$2 filters and 3-layer MLP with 1000 hidden units. We double the number of filters on CIFAR100. For text datasets, we use one 1D CNN layer with 1024/200 filters for THUCNews/DBPedia respectively and the same MLP as in image datasets\footnote{\citet{pgma} use pre-trained feature extractors whose architectures are unclear. The feature classifier of PGMA is the same MLP as we use. As their feature extractors are pre-trained on large external dataset, we think our classifiers are even weaker than theirs.}. We also use the same fixed pre-trained word embedding as in~\citep{pgma}. For all GFR methods, the AC-GAN based generator is to generate the 1000d $\mathbf{h}$ and $G$, $D$ are both 3-layer MLP for all datasets. The generator are trained with WGAN-GP technique. For GIR and OWM+GIR, on image data, the generators are also AC-GAN and trained with WGAN-GP where $G$, $D$ have 3 deconvolution and convolution layer respectively. On text data, $G$ is a 1-layer LSTM and $D$ is a 1-layer 1D CNN. We make the number of parameters in $G$ and $D$ comparable for GIR and GFR. We set $\lambda_{GP}=10$ in all experiments. For SSL auxiliary task, we choose $K=4$ and the rotation degrees are $0^{\circ}, 90^{\circ}, 180^{\circ}, 270^{\circ}$ respectively. $\alpha$ is chosen from $\{0.5, 1, 2, 5\}$ according to validation. Due to space limit, more details about implementations are in Appendix.

\subsection{Main Results}
We display the final accuracies on the whole test datasets after all tasks are learned in Table~\ref{table2}. We also plot the test accuracies on all learned tasks after each task for several settings in Figure~\ref{figure2}. It should be pointed out that as the vocabulary size of DBPedia is too large to fit SeqGAN in one 1080Ti GPU, we cannot obtain the results of GIR and OWM+GIR on this dataset. However, our GFR methods are not restricted by the vocabulary size as we need not generate original text.

Table~\ref{table2} shows that OWM performs much better than other baselines including regularization methods and GR methods in all settings. MeRGAN performs well on relatively simple SVHN but degrades on CIFAR10/100 in which the images are more complex. DGM utilizes task-specific masks~\citep{hat} to maintain old knowledge of generator and is slightly worse than OWM on SVHN and CIFAR10. However, on CIFAR100 it performs bad even if we endeavoured to adjust the hyperparameters. Moreover, DGM even performs worse after training $T_1$ as shown in Figure~\ref{figure2}, which may result from unstable training attributed to hard attention masks when the number of class is large. GIR method also performs poorly and unstably on text data. This phenomenon verifies that training generator continually on complex data can hardly succeed. As a result, combining GIR with OWM has negative effects except in 2 tasks CIFAR100, where the generator has not been trained in CL manner therefore there is no catastrophic forgetting in generator.

We find proposed method OWM+GFR/OWM+SSL+GFR performs best in 7 of 8 settings among all methods without real data storage. OWM+GFR outperform OWM by a significant margin except in 2 tasks and 20 tasks CIFAR100. For 20 tasks scenario, long task sequence exacerbates changes of features. OWM+GFR can be improved with SSL auxiliary loss consistently. Among GFR methods, naive GFR suffers from almost complete forgetting on previous tasks and performs like EWC, which means \textit{GFR can hardly work alone}. Meanwhile, FD+GFR works much worse than our OWM based GFR methods. We think those results demonstrate \textit{OWM is critical to GFR} and \textit{proposed methods (GFR and SSL) can improve OWM substantially}.

In addition, our method is even stronger or comparable to the real data replay baseline iCaRL although it works very well on simple datasets THUCNews and DBPedia. However iCaRL is ineffective on CIFAR100 where large number of classes restricts the ability of iCaRL's nearest mean-of-exemplars classifier.  

\subsubsection{Analysis of Feature Stability}
\begin{figure}[ht]
\centering
\includegraphics[height=0.16\textwidth]{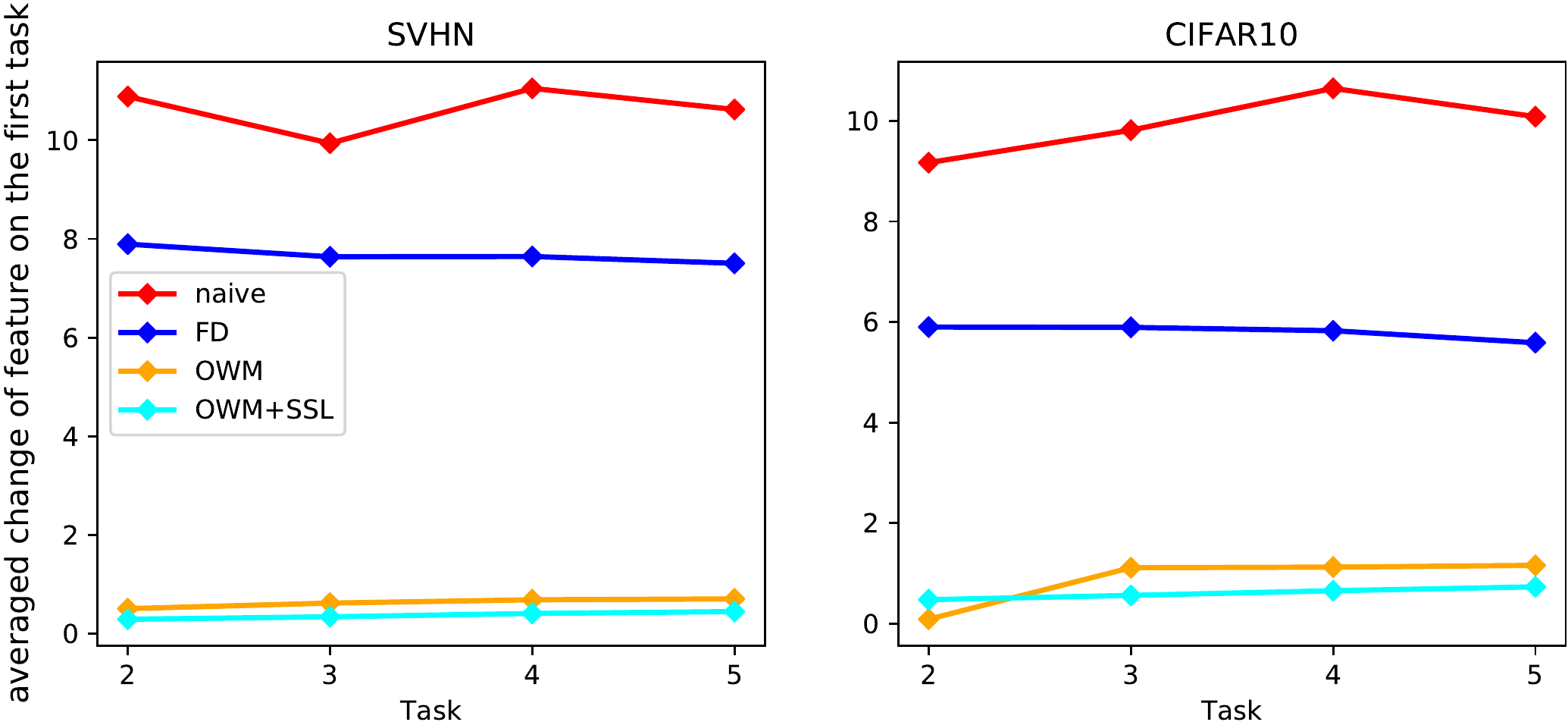}
\caption{The averaged $\Delta_i$ on different tasks.}
\label{figure4}
\end{figure}

To better understand how OWM and SSL auxiliary task are beneficial for the stability of features and why our methods are superior to FD+GFR~\citep{fdgfr}, we design an approach to measure the change of $\mathbf{h}$. After training $T_1$, we calculate the penultimate layer
feature and call it as $\mathbf{h}_{T_1}$. Then, after each subsequent task $T_i$ is trained, we calculate the new feature $\mathbf{h}_{T_i}$ and use $\Delta_i=\Vert\mathbf{h}_{T_1}-\mathbf{h}_{T_i}\Vert_2$ to evaluate the change of feature. We report the averaged $\Delta_i$ (where $i=2,3,4,5$) on the valid dataset of the first task of SVHN and CIFAR10 in Figure~\ref{figure4}. We find that compared to naive training without any constraint on feature, OWM can greatly eliminate the change of feature, while FD can only partially reduce the change. In addition, with SSL auxiliary task the change can be further reduced. These results explain why   
OWM is critical to the success of GFR and SSL auxiliary task helps enhance stability of feature $\mathbf{h}$.

\subsubsection{Error Analysis}
We delve into how GFR can improve OWM. An important deficiency of OWM is that data from different tasks are never trained together so that inferring classes from different tasks is potentially hard for classifier. We expect GFR can alleviate this problem. To verify this conjecture, we first divide the final classification error into two types: Inter-Task Error and Inner-Task Error. The former corresponds to task identity inference error while the latter refers to the classification error when task identity is inferred correctly. Formally they are defined as follows:
\begin{equation*}
\begin{split}
\mbox{Inter-Task Error} &= \frac{\sum_{i=1}^{|D_{test}|} \I[t_i\not=\hat{t}_i]}{|D_{test}|} \\
\mbox{Inner-Task Error} &= \frac{\sum_{i=1}^{|D_{test}|} \I[t_i=\hat{t}_i]\cdot \I[y_i\not=\hat{y}_i]}{|D_{test}|}
\end{split}
\end{equation*}
where $D_{test}$ is test dataset and $\I$ is the indicator function. $y_i$ and $\hat{y}_i$ are the true label and predicted label of $i$-th test sample respectively and $t_i$ and  $\hat{t}_i$ are respectively the task which $y_i$ and $\hat{y}_i$ belong to. We run OWM and our methods with the same random seed and calculate the above two types of error. The results of 3 settings are displayed in Table~\ref{table3}. 

\begin{table}[h]
\scriptsize

\begin{center}
\begin{tabular}{cccc}
\toprule[1.0pt]

Dataset&Method&Inter-Task Error(\%)&Inner-Task Error(\%)\\
 \hline \specialrule{0em}{1.0pt}{1.0pt} 
 
\multirow {3}{*}{SVHN} &OWM&21.59&4.58 \\ 
{} &OWM+GFR&20.12(-1.47)	&3.99(-0.59)\\ 
{} &OWM+SSL+GFR&17.87(-3.72)	&3.73(-0.85)\\ 
 \cline{1-4} \specialrule{0em}{0.pt}{0.pt} 
\multirow {3}{*}{CIFAR10} &OWM&41.52&3.83 \\ 
{} &OWM+GFR&38.72(-2.80)&4.87(+0.96) \\ 
{} &OWM+SSL+GFR&38.06(-3.46)	&4.07(+0.24)\\ 
\cline{1-4} \specialrule{0em}{0.5pt}{0.5pt} 
\multirow {3}{*}{\tabincell{c}{CIFAR100 \\(2 tasks)}} &OWM&27.65&30.06 \\ 
{} &OWM+GFR&29.31(+1.66)&28.15(-1.91) \\  
{} &OWM+SSL+GFR&27.90(+0.25)&28.70(-1.36) \\ 

\bottomrule[1.0pt] 
\end{tabular}
\end{center}
\caption{Comparison of two types of error between OWM and proposed methods in different settings. The gaps of each error are in parentheses.}
\label{table3}
\end{table}

Table~\ref{table3} shows in SVHN and CIFAR10, Inter-Task Error dominates the classification error of OWM. Furthermore, the improvement of OWM+GFR and OWM+SSL+GFR over OWM is dominated by that on Inter-Task Error. It should be noticed that Table~\ref{table2} shows on these two datasets, our methods outperform OWM significantly. We think these results can verify the conjecture that GFR can improve OWM by reducing Inter-Task Error. In contrast, in 2 tasks CIFAR100 setting, each task has 50 classes and Inner-Task Error has a similar level with Inter-Task Error. When applying GFR, the classifier tends to reduce Inner-Task Error which makes OWM+GFR performs similar with OWM. However, introducing SSL auxiliary task can reduce Inter-Task Error compared to OWM+GFR thus obtain better result.   

\begin{figure}[htb]
\centering
\subfloat[SVHN]{\label{fig3:a} \includegraphics[height=0.14\textwidth]{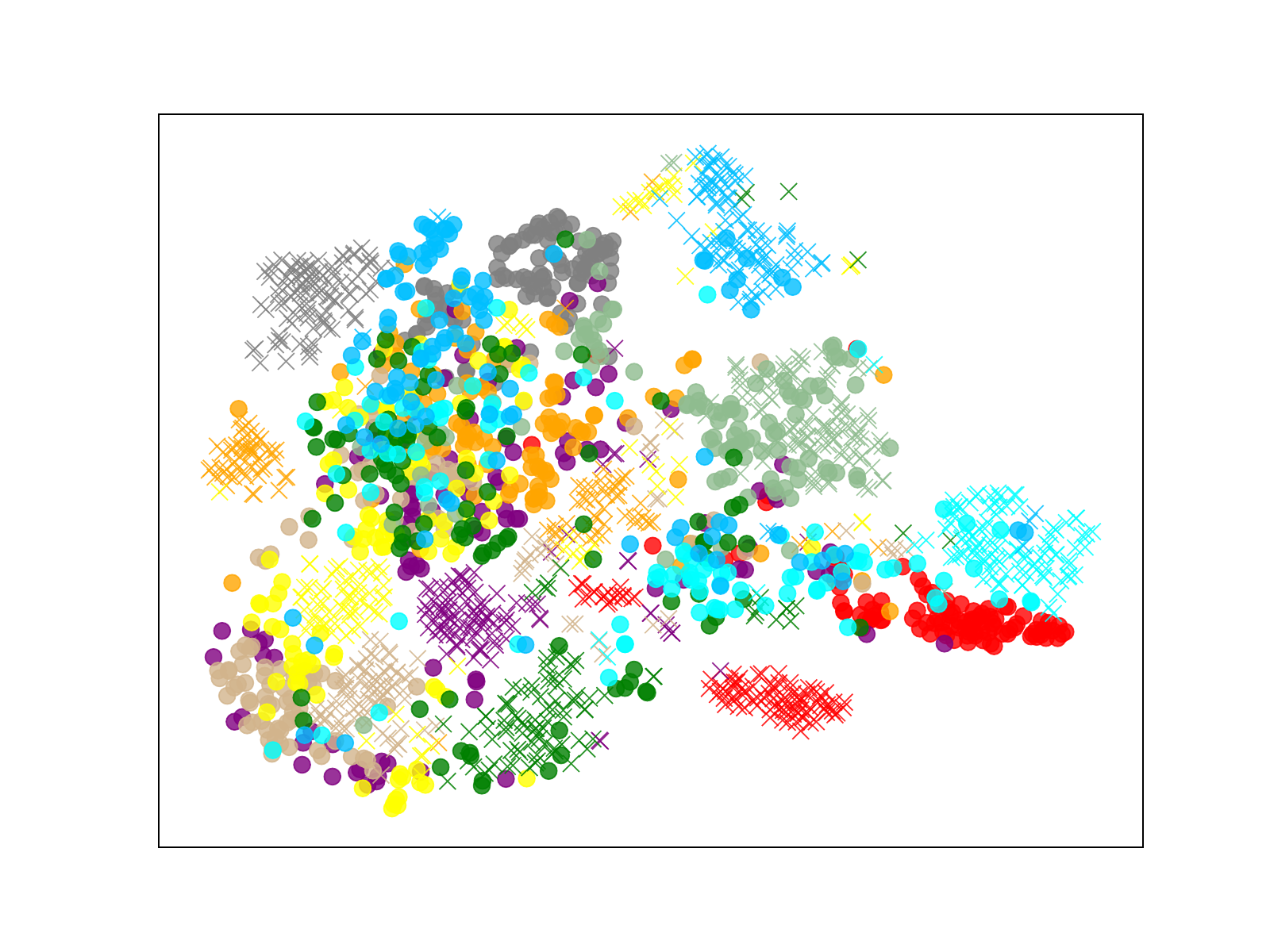}}\hspace{0mm}
	\subfloat[CIFAR10]{\label{fig3:b} \includegraphics[height=0.14\textwidth]{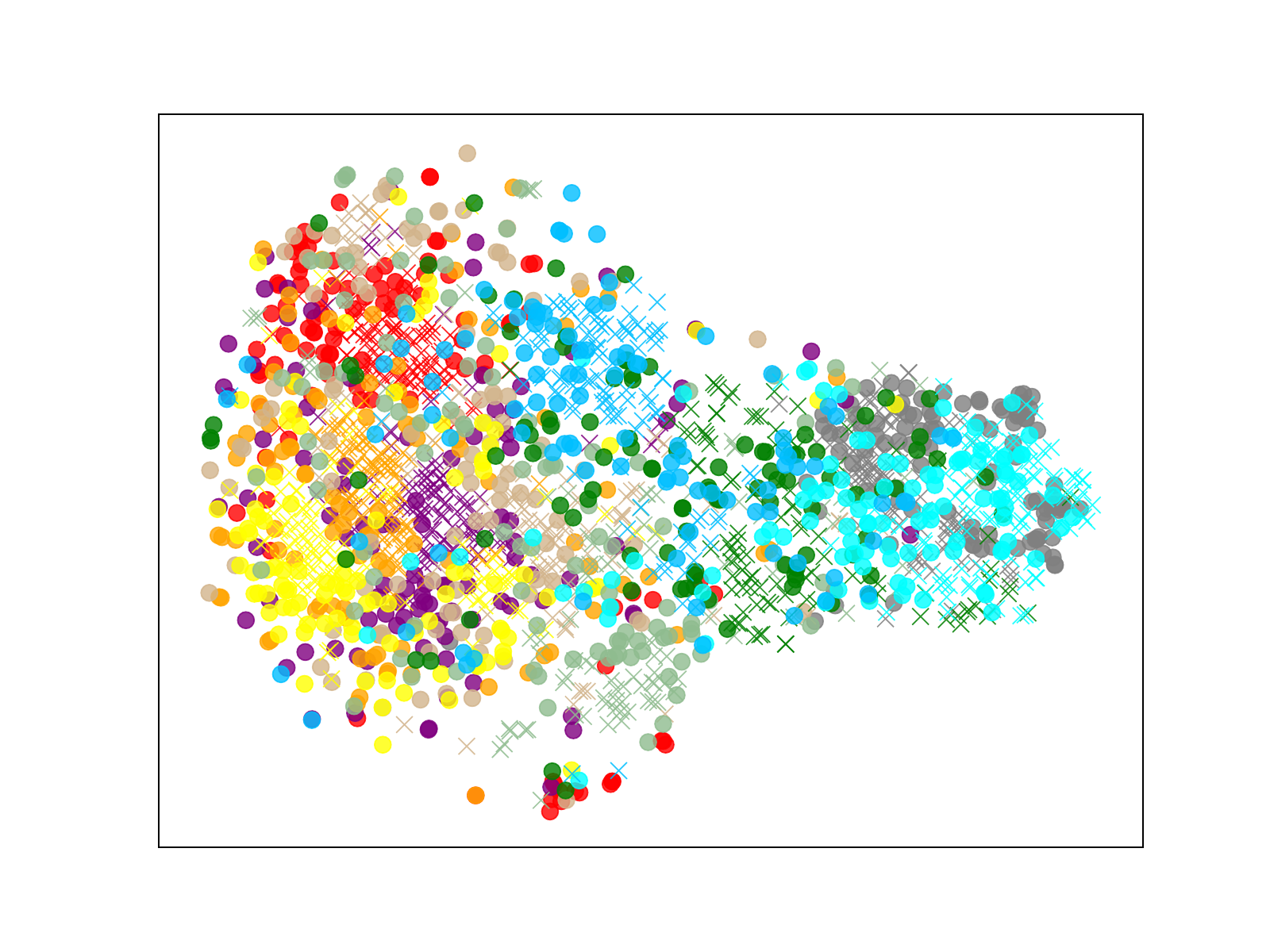}}
\caption{The visualization of real features ($\circ$) and fake features ($\times$) using t-SNE on SVHN and CIFAR10 datasets.}
\label{figure3}
\end{figure}

\subsection{Visualization of Features}
We visualize the generated penultimate layer features $\mathbf{h}$ as well as the real features to further explain why GFR is effective. For real features, we randomly select 100 samples in training dataset for each class and encode them using the final feature extractor $E$. We also randomly sample 100 samples for each class from the final generator $G$. We project all features to a joint 2D space using t-SNE~\citep{tsne}. The visualization results are in Figure~\ref{figure3}. We only plot the features on CIFAR10 and SVHN datasets due to space limit. We can observe a large part of generated features are clustered near some real feature clusters from the same class. Therefore, the generated features can provide useful information for classifier to adjust the decision surfaces for all classes simultaneously. We also observe some generated feature clusters, such as in grey and red, are far from corresponding real feature clusters. We find these classes are from the first or second task thus this phenomenon should attribute to the catastrophic forgetting of generator. 

\section{Conclusion}
In this paper, we focus on class incremental learning, a challenging CL scenario. With the help of OWM algorithm, we propose to GFR instead of convention GR to alleviate catastrophic forgetting. We also propose to employ an SSL auxiliary task to improve GFR based on OWM. We conduct experiments to not only demonstrate our method is superior to several strong baselines and achieves state-of-the-art performance without real data storage, but also discover that OWM is critical to GFR and why our method can improve OWM.
\bibliography{CLFR}

\end{document}